\PassOptionsToPackage{table}{xcolor}
\documentclass{article} % For LaTeX2e
\usepackage{iclr2027_conference,times}

\usepackage{amsmath,amsfonts,bm}

\def\eqref#1{equation~\ref{#1}}
\def\1{\bm{1}}

\DeclareMathAlphabet{\mathsfit}{\encodingdefault}{\sfdefault}{m}{sl}
\SetMathAlphabet{\mathsfit}{bold}{\encodingdefault}{\sfdefault}{bx}{n}

\usepackage{hyperref}
\usepackage{url}

\usepackage{comment}

\usepackage{amssymb}
\usepackage{booktabs}
\usepackage{multirow}
\usepackage{makecell}
\usepackage{graphicx}
\usepackage{wrapfig}
\usepackage{needspace}
\usepackage{fvextra}
\newif\ifsectionfourreview
\sectionfourreviewtrue

\usepackage{enumitem}
\usepackage{xcolor}
\newif\ifproofreadingreview
\proofreadingreviewtrue

\title{Can We Trust This Evaluation?\\
Instance-Level Reliability of LLM-as-a-Judge}

\author{
Yang Ba$^{1}$, Michelle V. Mancenido$^{2}$ \& Rong Pan$^{1}$\\
$^{1}$School of Computing and Augmented Intelligence, Arizona State University\\
$^{2}$School of Mathematical and Natural Sciences, Arizona State University\\
\texttt{\{yangba,mmanceni,Rong.Pan\}@asu.edu}
}

\iclrfinalcopy

\fancypagestyle{firstpagefooter}{
    \fancyhf{}
    \fancyfoot[L]{Preprint. Under review.}
    \renewcommand{\headrulewidth}{0pt}
    \renewcommand{\footrulewidth}{0pt}
}

\begin{document}

\maketitle
\thispagestyle{firstpagefooter}

\begin{abstract}
LLM-as-a-judge evaluation commonly aggregates multiple judges' decisions into a majority vote, yet this panel decision process does not guarantee that the evaluation of an individual item is reliable. We propose Structured Cross-fitted Uncertainty Risk (SCUR), a label-free approach for estimating instance-level evaluation reliability based on the judge panel's behavior. A latent measurement model serves as the backbone for measurement system analysis, from which SCUR derives three complementary signals capturing panel uncertainty, measurement cross-entropy, and perturbation sensitivity. These three sources of evidence are combined into a risk score. We evaluate the SCUR approach on five real-world benchmarks using majority-vote error ranking, selective trust, and risk stratification. Furthermore, we show our approach to be more effective at detecting high-consensus risk items, which are the blind spots for many other LLM-as-a-judge evaluation methods. The code is available at https://github.com/YangBa78/Evaluation-the-reliability-of-LLM-as-a-Judge. 

\end{abstract}

\section{Introduction}

\emph{LLM-as-a-judge} evaluation has become a standard approach for assessing open-ended generation tasks, including dialogue, reasoning, summarization, coding, instruction following, and safety \citep{liu2023geval,lin2023llmeval,li2025generation}, because it is scalable, inexpensive, and easy to deploy.
However, since the judge is itself a large language model (LLM), it can be biased, unstable, prompt-sensitive, and overconfident. A natural solution is to use multiple LLM judges to form a judge panel~\citep{zheng2023judging,verga2024replacing,chan2023chateval} and output only the majority-vote decision. A typical pipeline asks multiple LLMs to select one answer from two options---option $A$ or $B$---for a single question or prompt, then aggregates their votes into a label or score. This raises a fundamental question: \emph{How can we trust this panel's decision?}

Prior work has documented the option-display-position bias in LLM evaluators (i.e., the evaluator prefers a certain position, such as the first position, so leaning to A, instead of the real contents in these options~\citep{wang2023notfair,shi2024judging}), self- and model-family preference (i.e., different LLMs from the same foundation model family tend to vote similarly~\citep{wataoka2024self,li2025preference}), and broader instability in LLM judgments \citep{schroeder2024trust,li2025generation}. These concerns cannot be fully answered by simply aggregating multiple LLM judges' decisions \citep{zhang2023wider,chan2023chateval}. Note that these LLM evaluators may share overlapping training corpora, similar RLHF preferences, internet writing norms, safety alignment policies, benchmark contamination, and stylistic priors. As a result, \emph{multiple LLM judges may agree for the same wrong reason}. Thus, implicitly treating panel agreement as the evidence of answer correctness is very problematic \citep{wang2023notfair,shi2024judging, li2025preference,schroeder2024trust}. It may create a failure mode which we denote as \emph{``correlation collapse''}: a panel appears to be highly confident because most LLM judges agree with each other, but this agreement is driven by the shared judge bias rather than the underlying answer quality. In such cases, high consensus may produce false confidence, instead of a trustworthy decision. The existing LLM-evaluation aggregation methods focus on producing better decisions by weighting on each judge's reliability. In contrast, we study a complementary problem: given a panel's decision on one particular item, can we estimate how much this decision can be trusted?

\begin{figure*}[!t]
  \centering
  \includegraphics[width=\textwidth]{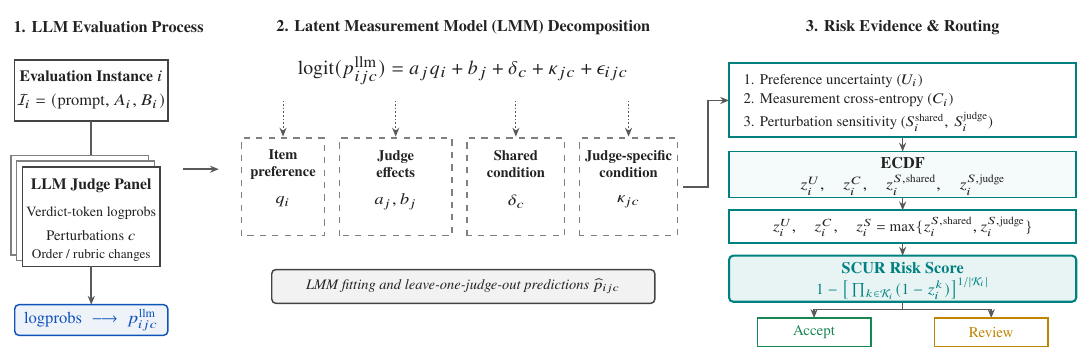}
  \caption{
  Overview of the proposed SCUR framework. The latent measurement model (LMM) decomposes LLM verdict probabilities into item, judge, and perturbation effects. Cross-fitted predictions yield preference uncertainty ($U$), measurement cross-entropy ($C$), and shared and judge-specific perturbation sensitivities. Each raw signal is separately normalized using its unlabeled reference ECDF; the two sensitivity percentiles are then combined by their maximum to obtain $z^S$. The available evidence $z^U,z^C,z^S$ is fused into SCUR for automatic acceptance or human review.}
  \label{fig:workflow}
\end{figure*}

Estimating the reliability of an evaluation panel's decision requires more than measuring the degree of panel agreement alone. We introduce a latent measurement model that employs cross-fitted, judge-level predictive probabilities under original and perturbation conditions. From these predictions, the Structured Cross-fitted Uncertainty Risk (SCUR) score is formulated, which preserves three interpretable misjudgment mechanisms---the latent uncertainty among judges, the alignment of predictive vs. native preference probabilities, and the panel's sensitivity to perturbation conditions---and combines their reference percentiles into a label-free risk score via a complement-based geometric mean.

We evaluate SCUR on five real-world LLM-as-a-judge benchmarks spanning pairwise preference, objective correctness, and instruction following. These experiments examine whether the proposed risk score correctly identifies unreliable panel decisions, improves evaluation reliability through selective review, and produces meaningful risk stratification across evaluated instances.

Our primary contributions are as follows:
\begin{itemize}[leftmargin=*]

\item We propose SCUR, a label-free approach to estimating decision reliability from a LLM-as-a-judge panel, without requiring ground-truth labels. It treats uncertainty, misaligned responses, and condition sensitivity as potential failure alarms for constructing a sensible risk score. 

\item We introduce an instance-level diagnostic analysis that decomposes the estimated risk into interpretable behavioral factors, providing insights into why a panel decision is deemed unreliable, then subsequently recommending human intervention.

\item We conduct extensive experiments on five real-world LLM-as-a-judge benchmarks, covering pairwise preference, objective correctness, and instruction following, and demonstrate that SCUR consistently outperforms existing uncertainty-only baselines in terms of majority-vote error ranking, selective trust, and risk stratification. Furthermore, we demonstrate that SCUR is effective for identifying high-consensus risky items and transferable to other vote aggregation rules.

\end{itemize}

\section{Related Work}

\paragraph{LLM-as-a-judge evaluation.}
LLM-as-a-judge methods use strong language models to evaluate human- or machine-generated
content through pairwise comparison, scalar scoring, or rubric-guided critique
\citep{lin2023llmeval,wang2023pandalm,kim2023prometheus}. These methods are attractive because
they scale beyond manual annotation, but they are not neutral measurement
devices. Prior work documents several deficiencies including position bias, verbosity bias, self-enhancement,
prompt sensitivity, and other evaluator artifacts
\citep{zheng2023judging,wang2023notfair,shi2024judging}. Recent work also
studies multi-judge or multi-role evaluator networks as a way to improve
alignment with human judgments \citep{zhang2023wider}. Our work is closest in
spirit to these reliability concerns, but shifts the target from average judge
quality to the reliability of each panel decision on individual instances.

\paragraph{Noisy annotators and label aggregation.}
The proposed framework is related to a long line of noisy-label aggregation and crowdsourcing models. Dawid--Skene estimates latent labels together with annotator confusion matrices from redundant noisy annotations \citep{dawid1979maximum}. Subsequent methods, including GLAD \citep{NIPS2009_f899139d}, MACE \citep{hovy2013learning}, and Bayesian annotation models \citep{paun2018comparing}, jointly model annotator competence, item difficulty, and annotation uncertainty to improve consensus estimation. These methods are designed to recover a more accurate latent label or aggregate decision. Recent work has improved LLM evaluation through confounder-aware panel aggregation \citep{zhao2026care}, item-response-theoretic modeling of judge behavior \citep{choi2026diagnosing}, and selective confidence estimation for pairwise judging \citep{badshah2026scope}.
In contrast, we assume that a panel decision has already been produced and address a complementary problem: estimating the reliability of that individual panel decision without requiring ground-truth labels.

\paragraph{Pairwise preference modeling.}
Pairwise evaluation is related to Bradley-Terry, Luce-choice, and related ranking models \citep{bradley1952rank,luce1959individual}. These models infer latent preference or quality scores from comparisons. In our setting, the comparison is contaminated by the evaluator bias and random noise due to the stochastic nature of generative LLMs, so the observed vote cannot be treated as deterministic. We model the LLM judge process itself and use the fitted structure to quantify the trustworthiness of the resulting panel decision.

\paragraph{Measurement system reliability.}
This work is also related to measurement system analysis, especially gauge repeatability and reproducibility, which partition the total
variation in measurements into variance components due to items, the measurement system, and repeated measurements~\citep{aiag2010msa}. A related line of work studies annotation quality in crowdsourced data through consistency, annotator credibility, variance decomposition, and spam diagnostics \citep{ba2025dataquality}. We adapt this perspective to LLM evaluation by treating items, judges, and perturbations as sources of measurement variation.

\section{Methodology}

We first introduce a latent measurement model (LMM) to model LLM-judge decisions under various designed perturbation schemes.  We then extract three failure mechanism-specific risk signals for the judge panel---decision uncertainty, measurement cross-entropy, and perturbation sensitivity---and combine their reference percentiles into a risk-ranking score.

\subsection{Latent Measurement Model}

Suppose there are two options for an LLM judge to choose from for each piece of item to be evaluated. Each item consists of a prompt, two candidate answers, and an optional judging rubric, and it is denoted by
$\mathcal I_i=(\mathrm{prompt}_i,A_i,B_i,\mathrm{rubric}_i).$
For item \(i\), judge \(j\), and perturbation condition \(c\in\mathcal C\), let \(p^{\mathrm{llm}}_{ijc}\)\(\in(0,1)\) be the judge's response probability that the \emph{canonical} candidate \(A_i\) wins. \(p^{\mathrm{llm}}_{ijc}\) is obtained by exponentiating and normalizing the judge's verdict-token logprobs, the logarithms of the probabilities the LLM assigns to the possible verdicts.  We reserve \(c=0\) for the original condition: the unperturbed answer order, prompt, and rubric.  When \(c\neq0\) it denotes one prespecified intervention, such as swapping the display order of answers or changing the judging rubric.  After an answer-order swap, for example, the probability is mapped back to the same canonical event ``original \(A_i\) wins,'' so probabilities remain comparable between the original and perturbed conditions.

\(p^{\mathrm{llm}}_{ijc}\) has been widely used to measure how confident a large language model is in the verdict token it generates. We employ a latent measurement model (LMM) to decompose its log-odds, making item, judge, and condition effects separable:
\begin{equation}
\label{eq:soft-hlm}
\operatorname{logit}\!\left(p^{\mathrm{llm}}_{ijc}\right)
=a_jq_i+b_j+\delta_c+\kappa_{jc}+\epsilon_{ijc},
\qquad \epsilon_{ijc}\sim\mathcal N(0,\sigma_j^2).
\end{equation}
This decomposition distinguishes the item-level preference \(q_i\) from systematic judge and condition effects and residual variation, allowing individual judge responses to be assessed relative to their model-implied preferences.
The latent variable \(q_i\) represents the panel's underlying consensus preference for the canonical candidate, inferred jointly from all judges' responses rather than a correctness oracle. The discrimination \(a_j=\exp(\alpha_j)>0\) controls how strongly judge \(j\) responds to that consensus, as in a two-parameter item-response model \citep{lord1980applications}; \(b_j\) is this judge's overall tendency toward canonical candidate \(A\). The shared perturbation effect \(\delta_c\) captures the panel-wide response to intervention \(c\), whereas \(\kappa_{jc}\) captures judge \(j\)'s deviation from that shared response.  We set \(\delta_0=\kappa_{j0}=0\), so the unperturbed original item serves as a reference.

\paragraph{Estimation.} We estimate the LMM mean structure using alternating penalized least-squares updates with equal observation weights. The quadratic penalties admit a Gaussian-shrinkage interpretation, shrinking \(q_i,b_j,\delta_c,\kappa_{jc}\) toward zero and \(a_j\) toward one. In particular, shrinking \(\kappa_{jc}\) toward zero regularizes each judge’s condition effect toward the shared effect \(\delta_c\). Judge-specific residual variances are estimated after fitting the mean structure. Appendix~\ref{app:model-structure} provides the penalty strengths, scale and centering conventions, and optimization details.

\paragraph{Dataset partition and cross-fitting.}
We partition each benchmark by items in a 50/30/20 ratio into \(\mathcal D_{\mathrm{fit}}/\mathcal D_{\mathrm{ref}}/\mathcal D_{\mathrm{eval}}\) datasets:
\(\mathcal D_{\mathrm{fit}}\) estimates LMM parameters,
\(\mathcal D_{\mathrm{ref}}\) supplies the reference distributions for risk normalization,
and \(\mathcal D_{\mathrm{eval}}\) is held out for evaluation.
No panel decision-error labels are used for LMM fitting or risk estimation.
For each reference or evaluation item, we freeze the fitted judge and
condition parameters and infer \(q_i^{(-j)}\) from other judges,
excluding judge \(j\)'s responses for that item. The cross-fitted
prediction is \(\widehat p_{ijc}^{(-j)}
=\sigma(a_jq_i^{(-j)}+b_j+\delta_c+\kappa_{jc})\), where \(\sigma\)
is the logistic function; for simplicity, we rewrite it as \(\widehat p_{ijc}\).
Excluding judge \(j\)'s own responses ensures that this prediction is based on evidence from the remaining panel, preventing a prediction contamination due to self-response reinforcement.

\subsection{Evidence-Based Risk Signals}

We define three complementary risk signals: \(U\) measures ambiguity in
the pooled preference, \(C\) assesses individual judge responses against
the learned panel preference structure, and \(S\) measures dependence on perturbation conditions.

\paragraph{Model-based preference uncertainty.}

Let \(\overline p_i=J_i^{-1}\sum_{j=1}^{J_i}\widehat p_{ij0}\), where
\(J_i\) is the number of available judges for item \(i\), be the mean
cross-fitted preference probability under the original condition. We define
\[
\label{eq:risk-u}
U_i=H(\overline p_i),
\qquad
H(p)=-p\log p-(1-p)\log(1-p).
\]
\(U_i\) is largest when \(\overline p_i\) is close to 0.5, meaning that the panel does not clearly favor either candidate. This may occur because individual judges are uncertain or because different judges confidently favor different candidates. However, it may remain low when the judges confidently agree and therefore cannot identify errors shared across the panel.

\paragraph{Measurement cross-entropy.}

We compare each judge's LLM confidence \(p^{\mathrm{llm}}_{ij0}\) with its
cross-fitted LMM prediction \(\widehat p_{ij0}\) under the original condition
using binary cross-entropy:
$$
C_i=-\frac{1}{J_i}\sum_{j=1}^{J_i}
\left[
p^{\mathrm{llm}}_{ij0}\log\widehat p_{ij0}
+(1-p^{\mathrm{llm}}_{ij0})\log(1-\widehat p_{ij0})
\right].
$$
\(C_i\) increases when a judge's LLM confidence disagrees with the
cross-fitted prediction based on the remaining panel. The mismatch is
greater when the judge confidently favors one candidate while the
cross-fitted prediction favors the other. Such a mismatch indicates that
the judge's confidence is not well supported by the preference structure
inferred from the remaining panel, making it informative for assessing
decision risk. Cross-entropy also reflects the uncertainty of the LLM
confidence itself, so \(C_i\) can be high even when the two probabilities are similar but uncertain.
Thus, \(C_i\) reflects both individual-judge uncertainty and cross-fitted
mismatch, complementing the pooled preference uncertainty measured by \(U_i\).

\paragraph{Perturbation sensitivity.}
For benchmarks with paired perturbations, where the same item is evaluated under the original and perturbed conditions, let \(\mathcal C_i^+=\mathcal C_i\setminus\{0\}\) denote the available perturbed conditions, such as answer-order swaps, strict-rubric changes, and evidence-first rubric changes. Let \(\eta_{ij0}=a_jq_i^{(-j)}+b_j\). Keeping the cross-fitted item margin fixed, we separately measure the probability changes due to the shared and judge-specific condition effects:
\begin{align*}
S_i^{\mathrm{shared}}
&=\frac{1}{J_i|\mathcal C_i^+|}
\sum_{j=1}^{J_i}\sum_{c\in\mathcal C_i^+}
\left|\sigma(\eta_{ij0}+\delta_c)-\sigma(\eta_{ij0})\right|,\\
\label{eq:risk-s-judge}
S_i^{\mathrm{judge}}
&=\frac{1}{J_i|\mathcal C_i^+|}
\sum_{j=1}^{J_i}\sum_{c\in\mathcal C_i^+}
\left|\sigma(\eta_{ij0}+\delta_c+\kappa_{jc})
-\sigma(\eta_{ij0}+\delta_c)\right|.
\end{align*}
\(S_i^{\mathrm{shared}}\) measures the probability change associated with a perturbation across the panel, while \(S_i^{\mathrm{judge}}\) measures additional judge-specific changes beyond this shared effect. 
These two components capture complementary forms of perturbation sensitivity and are combined after reference normalization, as described below. Large changes indicate that a preference depends on the judging conditions, which can undermine its stability.

\subsection{Label-Free Risk Score}

\paragraph{Reference normalization.}
We use \(r_i^k\) for raw signal \(k\) and \(z_i^k\) for its normalized
evidence.  Specifically, let
\(r_i^U=U_i\), \(r_i^C=C_i\), \(r_i^{S,\mathrm{shared}} = S_i^{\mathrm{shared}}\),
and
\(r_i^{S,\mathrm{judge}} = S_i^{\mathrm{judge}}\). For each available raw signal \(k\), we
estimate an empirical cumulative distribution function (ECDF)
\citep{wasserman2004all} on the unlabeled reference split:
\[
\label{eq:ecdf-evidence}
z_i^k
=\widehat F_{k,\mathcal D_{\mathrm{ref}}}(r_i^k)
=\frac{1}{|\mathcal D_{\mathrm{ref}}^k|}
\sum_{h\in\mathcal D_{\mathrm{ref}}^k}
\mathbf 1\!\left\{r_h^k\le r_i^k\right\},
\]
where \(\mathcal D_{\mathrm{ref}}^k\) contains reference items with risk signal \(k\).  The ECDF is estimated without golden (true) labels, and it is then frozen for evaluation. The transformation from $r_i^k$ to $z_i^k$ maps heterogeneous raw signal scales to comparable
percentiles.  
For perturbation sensitivity, we then combine its two components as
\[
z_i^S
=
\max\left\{
z_i^{S,\mathrm{shared}},
z_i^{S,\mathrm{judge}}
\right\}.
\]
Taking the maximum after normalization retains whichever form of perturbation sensitivity is more unusual relative to its reference distribution. Therefore, the normalized evidence captures three complementary aspects of decision risk: preference ambiguity (\(z_i^U\)), individual-judge uncertainty and mismatch (\(z_i^C\)), and perturbation sensitivity (\(z_i^S\)).

\paragraph{Structured Cross-fitted Uncertainty Risk.}

Let \(\mathcal K_i\) contain the available channels: \(\{U,C,S\}\) with
paired perturbations and \(\{U,C\}\) otherwise.
We construct \emph{Structured Cross-fitted Uncertainty Risk} (\(\mathrm{SCUR}\)) as the
complement of the geometric mean of the complements of normalized evidence:
\begin{equation}
\label{eq:final-direct-scur}
\mathrm{SCUR}_i
=
1-
\left[
\prod_{k\in\mathcal K_i}(1-z_i^k)
\right]^{1/|\mathcal K_i|}.
\end{equation}
Specifically, \(\mathrm{SCUR}_i =
1-[(1-z_i^U)(1-z_i^C)(1-z_i^S)]^{1/3}\).
The complement-based geometric mean aggregates the available risk evidence:
a high percentile for any evidence type reduces its complement and increases
the resulting SCUR score. The \(|\mathcal K_i|\)-th root accounts for the
number of available evidence types, allowing the same aggregation rule when
perturbation sensitivity is unavailable. The resulting
\(\mathrm{SCUR}_i\in[0,1]\) provides a relative risk score, which will be used to rank items
for automatic acceptance or human review.

\section{Experiments}
\label{sec:experiment}

\newcommand{\rawbaselinefigure}[2]{%
\ifsectionfourreview
{\setlength{\fboxsep}{2pt}\setlength{\fboxrule}{.6pt}%
\begin{minipage}{\linewidth}\centering
{\color{red}\scriptsize Previous}\par\nobreak
{\color{red}\fbox{\begin{minipage}{\dimexpr\linewidth-2\fboxsep-2\fboxrule\relax}\centering#1\end{minipage}}}\par
\vspace{3pt}
{\color{blue}\scriptsize Updated}\par\nobreak
{\color{blue}\fbox{\begin{minipage}{\dimexpr\linewidth-2\fboxsep-2\fboxrule\relax}\centering#2\end{minipage}}}
\end{minipage}}%
\else#2\fi}

\subsection{Benchmark Evaluation}
\label{sec:real}

\begin{table*}[!t]
\caption{
\textbf{Majority-vote error ranking} (Mean $\pm$ standard deviation over five independent trials). Panel Err. is the original-condition majority-vote error rate. Err@50 is the error rate among the lowest-risk 50\% of evaluation items. SCUR outperforms baselines on each benchmark.}
\label{tab:summary}
\vspace{0.5em}

\centering
\scriptsize
\resizebox{\textwidth}{!}{%
\begin{tabular}{l l c c l c c c}
\toprule
Benchmark & Sample Sizes & \#Judges & Panel Err. & Method
& AUROC $\uparrow$ & $1-\mathrm{AURC}$ $\uparrow$ & Err@50 $\downarrow$ \\
\midrule

\multirow{5}{*}{Chatbot-Arena}
& \multirow{5}{*}{\makecell[l]{
\#Obs.: 589,050\\
\#Items: 17,325\\
\#Ref.: 5,198\\
\#Test: 3,465
}}
& \multirow{5}{*}{17}
& \multirow{5}{*}{.3427 $\pm$ .0082}
& PE
& .674 $\pm$ .008
& .773 $\pm$ .010
& .229 $\pm$ .014 \\

& & & & VE
& .625 $\pm$ .007
& .741 $\pm$ .008
& .252 $\pm$ .011 \\

& & & & DS entropy
& .673 $\pm$ .006
& .765 $\pm$ .009
& .222 $\pm$ .010 \\

& & & & MSP
& .647 $\pm$ .011
& .762 $\pm$ .012
& .241 $\pm$ .011 \\

& & & & \cellcolor{gray!15}\textbf{SCUR}
& \cellcolor{gray!15}.696 $\pm$ .007
& \cellcolor{gray!15}.796 $\pm$ .008
& \cellcolor{gray!15}.195 $\pm$ .008 \\

\midrule

\multirow{5}{*}{HaluEval}
& \multirow{5}{*}{\makecell[l]{
\#Obs.: 120,000\\
\#Items: 20,000\\
\#Ref.: 6,000\\
\#Test: 4,000
}}
& \multirow{5}{*}{6}
& \multirow{5}{*}{.1937 $\pm$ .0035}
& PE
& .728 $\pm$ .006
& .891 $\pm$ .004
& .081 $\pm$ .005 \\

& & & & VE
& .667 $\pm$ .003
& .868 $\pm$ .004
& .129 $\pm$ .003 \\

& & & & DS entropy
& .674 $\pm$ .004
& .868 $\pm$ .004
& .122 $\pm$ .003 \\

& & & & MSP
& .718 $\pm$ .011
& .897 $\pm$ .005
& .093 $\pm$ .006 \\

& & & &
\cellcolor{gray!15}\textbf{SCUR}
& \cellcolor{gray!15}.732 $\pm$ .006
& \cellcolor{gray!15}.901 $\pm$ .005
& \cellcolor{gray!15}.079 $\pm$ .005 \\

\midrule

\multirow{5}{*}{MMLU}
& \multirow{5}{*}{\makecell[l]{
\#Obs.: 477,428\\
\#Items: 28,084\\
\#Ref.: 8,425\\
\#Test: 5,617
}}
& \multirow{5}{*}{17}
& \multirow{5}{*}{.2295 $\pm$ .0046}
& PE
& .787 $\pm$ .004
& .907 $\pm$ .003
& .079 $\pm$ .003 \\

& & & & VE
& .744 $\pm$ .005
& .892 $\pm$ .003
& .098 $\pm$ .005 \\

& & & & DS entropy
& .730 $\pm$ .007
& .886 $\pm$ .003
& .093 $\pm$ .003 \\

& & & & MSP
& .693 $\pm$ .006
& .878 $\pm$ .003
& .129 $\pm$ .005 \\

& & & &
\cellcolor{gray!15}\textbf{SCUR}
& \cellcolor{gray!15}.794 $\pm$ .005
& \cellcolor{gray!15}.915 $\pm$ .003
& \cellcolor{gray!15}.062 $\pm$ .003 \\

\midrule

\multirow{5}{*}{LLMBar}
& \multirow{5}{*}{\makecell[l]{
\#Obs.: 12,570\\
\#Items: 419\\
\#Ref.: 126\\
\#Test: 83
}}
& \multirow{5}{*}{5}
& \multirow{5}{*}{.3711 $\pm$ .0198}
& PE
& .623 $\pm$ .048
& \makecell{.711 $\pm$ .029}
& .290 $\pm$ .031 \\

& & & & VE
& .585 $\pm$ .046
& .677 $\pm$ .028
& .311 $\pm$ .034 \\

& & & & DS entropy
& .640 $\pm$ .030
& .704 $\pm$ .019
& .265 $\pm$ .013 \\

& & & & MSP
& .498 $\pm$ .050
& .651 $\pm$ .027
& .381 $\pm$ .029 \\

& & & &
\cellcolor{gray!15}\textbf{SCUR}
& \cellcolor{gray!15}.654 $\pm$ .042
& \cellcolor{gray!15}.732 $\pm$ .021
& \cellcolor{gray!15}.238 $\pm$ .017 \\

\midrule

\multirow{5}{*}{LLMEval\textsuperscript{2}}
& \multirow{5}{*}{\makecell[l]{
\#Obs.: 6,000\\
\#Items: 200\\
\#Ref.: 60\\
\#Test: 40
}}
& \multirow{5}{*}{5}
& \multirow{5}{*}{.2100 $\pm$ .0418}
& PE
& \makecell{.553 $\pm$ .068}
& \makecell{.810 $\pm$ .056}
& \makecell{.190 $\pm$ .055} \\

& & & & VE
& .523 $\pm$ .086
& .797 $\pm$ .055
& .197 $\pm$ .040 \\

& & & & DS entropy
& .546 $\pm$ .064
& .807 $\pm$ .034
& .200 $\pm$ .041 \\

& & & & MSP
& .603 $\pm$ .122
& .824 $\pm$ .077
& .170 $\pm$ .076 \\

& & & &
\cellcolor{gray!15}\textbf{SCUR}
& \cellcolor{gray!15}.632 $\pm$ .070
& \cellcolor{gray!15}.861 $\pm$ .036
& \cellcolor{gray!15}.160 $\pm$ .042 \\

\bottomrule
\end{tabular}%
}

\end{table*}

\begin{figure*}[!t]
\centering \includegraphics[width=\linewidth]{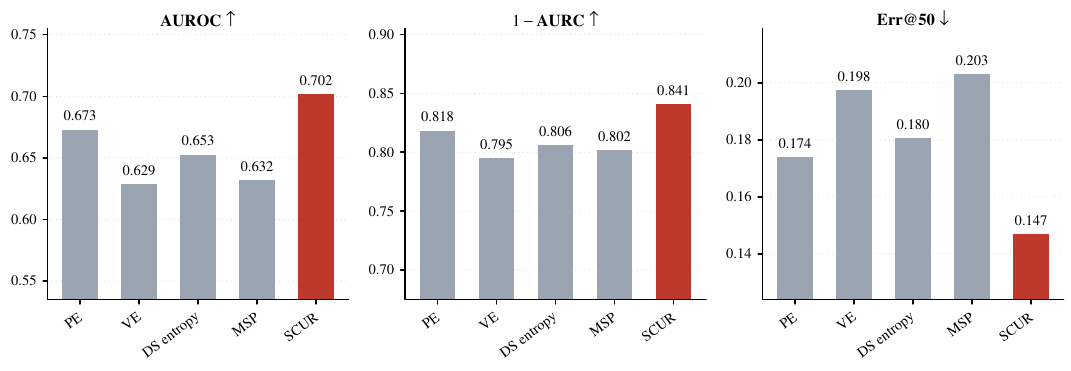}
\caption{\textbf{Average performance across five benchmarks. }
Each benchmark receives equal weight. SCUR (red) has the highest AUROC
and $1-\mathrm{AURC}$ and the lowest Err@50; gray bars show the four baselines.}
\label{fig:average-results}
\end{figure*}

\begin{figure*}[!htbp]
    \centering
    \includegraphics[width=.94\linewidth]{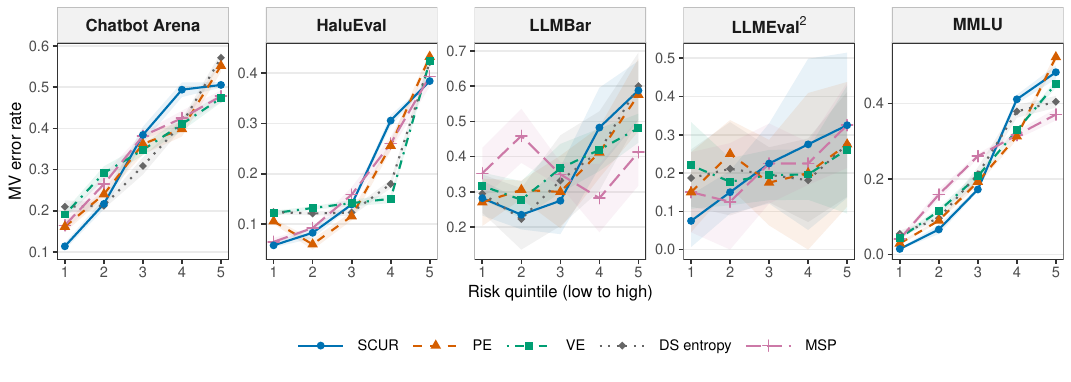}
    \includegraphics[width=.94\linewidth]{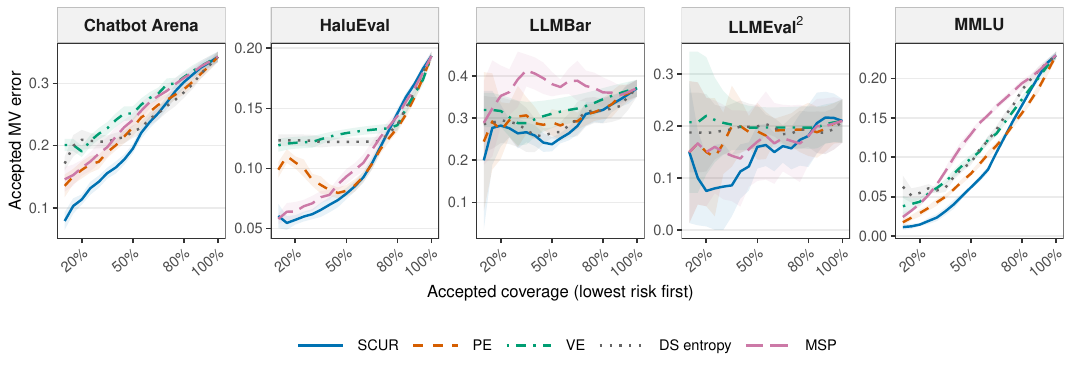}
    \caption{
    \textbf{Majority-vote risk stratification and selective trust.}
    Top: error across increasing risk quintiles; monotonic increases indicate that
    higher-risk quintiles contain more errors. Bottom: error among the lowest-risk
    items retained at each coverage; lower curves indicate fewer errors at the same
    coverage. Lines show five-seed means, with shaded bands indicating $\pm1$
    standard deviation.} 
    \label{fig:risk}
\end{figure*}

\paragraph{Benchmarks.} We validate the proposed reliability framework on five multi-LLM-as-a-judge benchmarks:
Chatbot-Arena \citep{zheng2023judging}, HaluEval \citep{li2023halueval},
\mbox{MMLU~\citep{hendrycks2021measuring}}, LLMBar
\citep{zeng2024evaluating,gao2024bayesian}, and LLMEval\textsuperscript{2}
\citep{zhang2023wider,gao2024bayesian,zeng2024evaluating}.  They cover
pairwise preference, hallucination detection, multiple-choice correctness,
instruction following, and multidimensional prompt-following. We use a 50/30/20 data split: 50\% fits the LMM, 30\% supplies the reference distributions for SCUR, and 20\%
is untouched test data.  We run 5 independent trials.  

Chatbot-Arena, HaluEval, and MMLU use existing judge outputs.
Chatbot-Arena includes original and swapped answer orders; HaluEval and
MMLU contain only the original condition. For LLMBar and LLMEval\textsuperscript{2}, we
collect verdict-token log-probabilities from five judge models, crossing
original/swapped answer order with standard, strict, and evidence-first
rubrics to obtain six conditions per item--judge pair.
Thus Chatbot-Arena, LLMBar, and LLMEval\textsuperscript{2} provide $U,C,S$, whereas HaluEval
and MMLU provide only $U,C$. See Appendix~\ref{app:perturbations} for the
benchmark-specific conditions and judging protocol.

\paragraph{Risk baselines.}
We compare SCUR with four panel-level uncertainty baselines: vote entropy
(VE), which is the binary entropy of the majority-vote rate related to vote-agreement
confidence used for LLM judges \citep{radharapu2025calibrating}; predictive
entropy (PE), which is the entropy of the panel-mean probability, extending
predictive entropy used for LLM-judge abstention \citep{badshah2026judge};
DS entropy, which is the posterior entropy of the Dawid-Skene model, adapting its
LLM-evaluator aggregation applications \citep{gao2024bayesian,zhao2026care}
to risk scoring; and mean per-judge MSP uncertainty
\citep{hendrycks2017baseline,radharapu2025calibrating}.
DS entropy ranks items by uncertainty; DS aggregation, discussed later in this section,
instead determines their verdicts. Appendix~\ref{app:risk-baselines}
specifies the formulas and the panel adaptations of these prior studies.

\paragraph{Evaluation metrics.}
Our primary target is the error made by majority voting verdict,
\(e_i^{\mathrm{MV}}=\mathbf1\{\widehat y_i^{\mathrm{MV}}\ne y_i^{\mathrm{gold}}\}\).
AUROC measures how well risk separates incorrect from correct verdicts;
we report $1-\mathrm{AURC}$, the complement of area under the risk--coverage
curve \citep{geifman2019bias,zhou2025aurc}, equivalently the mean retained
accuracy over all coverage levels $k/n$, $k=1,\ldots,n$ (higher is better);
and Err@50
is the error rate among the lowest-risk half of evaluation items.
We also examine error by increasing risk quintile and across retained
coverage. For an alternative aggregation rule $g$ (Section 4.3), we recompute the error as
\(e_i^{g}=\mathbf1\{\widehat y_i^{g}\ne y_i^{\mathrm{gold}}\}\), while using the same SCUR scores to rank the items.

\paragraph{Majority-vote error ranking.}
Table~\ref{tab:summary} shows that SCUR has the highest mean AUROC and $1-\mathrm{AURC}$
and the lowest mean Err@50 on each benchmark. Figure~\ref{fig:average-results} summarizes the equally weighted benchmark averages. Compared with PE, the best baseline on average, AUROC increases from .673 to .702, $1-\mathrm{AURC}$ from .818 to .841, and Err@50 decreases from .174 to .147. The improvement in AUROC shows that SCUR better separates incorrect
from correct panel decisions. The lower Err@50 gives this ranking gain an
operational meaning: at the same 50\% acceptance budget, the retained
set contains fewer errors. The $1-\mathrm{AURC}$ improvement shows that this advantage persists across coverage levels rather than at a single cutoff. Overall, SCUR provides a more effective ranking of panel-error risk than the uncertainty baselines.

\paragraph{\normalcolor Selective trust and risk strata.}

Figure~\ref{fig:risk} examines how observed error changes with SCUR risk and retained coverage. On Chatbot-Arena, HaluEval, and MMLU, error increases monotonically from the lowest to the highest SCUR quintile, showing that higher SCUR scores correspond to higher empirical error. LLMEval\textsuperscript{2} shows the same monotonic pattern, while LLMBar shows some variation among the lower-risk quintiles but substantially higher error in the two highest-risk groups.

Across the coverage curves, SCUR generally achieves lower error at the same coverage than the baselines, particularly at lower coverage. On LLMBar and LLMEval\textsuperscript{2}, which have substantially fewer evaluation items than the other benchmarks, estimates are more variable and the baselines show weak or inconsistent ordering across risk quintiles. Despite this increased variability, SCUR retains a clearer separation between lower- and higher-risk items.

\subsection{High-Consensus Blind Spots}
\begin{wrapfigure}{r}{0.51\textwidth}
\vspace{-9pt}
\centering\includegraphics[width=\linewidth]{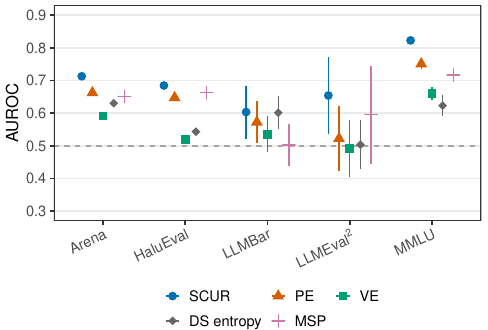}
\caption{\textbf{Performance on high-consensus items.}
AUROC on items with at least 80\% vote agreement.
Points show five-seed means and whiskers indicate $\pm1$ standard deviation.
The dashed line marks chance-level AUROC.}
\label{fig:blindspot-exact-pattern}
\vspace{-10pt}
\end{wrapfigure}

High-consensus (HC) errors are challenging cases for any disagreement-based risk measurement, because incorrect verdicts can remain hidden when most judges agree on this incorrect choice.  We therefore evaluate items for which at least 80\% of judges choose the same answer under the original condition, using each judge's binary A/B choice. This defines the high-consensus (HC) subset; gold labels are used only to determine whether the majority verdict is correct. We ask whether SCUR can still distinguish incorrect from correct verdicts by marking them with high risk.

Figure~\ref{fig:blindspot-exact-pattern} shows that SCUR achieves the highest
mean AUROC on the HC subset across all five benchmarks. The advantage is
particularly clear on the larger benchmarks, while estimates are more variable
on LLMBar and LLMEval\textsuperscript{2}. These results show that SCUR can identify risk differences even among high-consensus decisions, where other methods struggle to separate incorrect verdicts from correct verdicts.

The matched-vote analysis in Appendix~\ref{app:hc-supplement} provides a stricter test by comparing items with identical original-condition vote patterns across judges (e.g., a 4-to-1 vote is compared only with other 4-to-1 votes in which the same four judges support the same candidate). The appendix also considers an alternative definition of high consensus based on panel-mean probabilities, selecting items with a panel-mean confidence of at least .8. Across both settings, SCUR outperforms the baselines in separating incorrect from correct verdicts, indicating that SCUR consistently identifies panel errors that remain difficult to detect under high consensus.

\subsection{Risk Generalizability}

SCUR assesses the reliability of panel decisions independently of how judge evaluations are aggregated, allowing it to be applied to different aggregation rules. We evaluate this setting with Dawid--Skene (DS), MACE, GLAD, and CARE \citep{zhao2026care} in addition to majority vote. Each method produces its own verdict and error labels. 

\begin{figure*}[!t]
\centering
\includegraphics[width=.99\textwidth]{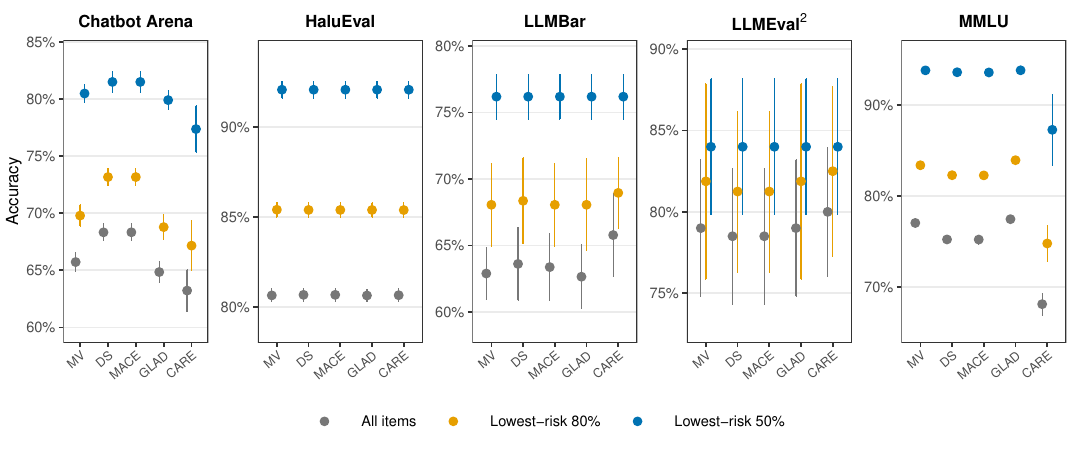}
\caption{
\textbf{Selective acceptance across five aggregation rules.} 
Gray: all evaluation items; orange/blue: lowest-risk 80\%/50\% based on SCUR ranking. Points and whiskers show five-seed means and $\pm1$
standard deviation; y-axis ranges vary by benchmark.
}
\label{fig:selective-acceptance-aggregation}
\end{figure*}

Figure~\ref{fig:selective-acceptance-aggregation} shows that retaining lower-risk decisions according to SCUR improves accuracy across all five benchmarks and five aggregation rules, with an average gain of 4.6 pp at 80\% lowest-risk threshold and 12.2 pp at 50\% threshold, compared with accepting all panel decisions. These results show that SCUR is robust to different aggregation methods; therefore, it can serve as a common risk-assessment metric for either accepting the panel decision or moving the case to human review.

\subsection{Panel-Decision Diagnosis}
\label{sec:panel-decision-diagnosis}

Table~\ref{tab:instance-diagnostics} illustrates how the three risk signals contribute to individual review decisions. High \(z^U\), \(z^C\), and \(z^S\) indicate preference ambiguity, judge-level mismatch, and sensitivity to perturbations, respectively; high \(z^C\) may also reflect uncertain LLM confidence. Together, the signals provide complementary evidence for why an item may warrant review without requiring a gold label.

The examples listed in the table show that elevated risk can arise under strong panel agreement, without perturbation evidence, or through several moderately elevated signals rather than a single dominant one. The last row (MMLU, Instance ID 19396) provides a low-risk contrast.

\begin{table*}[t]
\caption{
\textbf{Instance-level diagnostics on held-out items.}
Panel/Gold reports the majority verdict and gold label. Vote agreement is the
fraction of judges supporting the original-condition majority verdict.
The $z$ values are reference percentiles, and -- indicates that the
corresponding risk evidence is unavailable.
}
\vspace{0.5em}
\label{tab:instance-diagnostics}
\centering
\scriptsize
\setlength{\tabcolsep}{2.2pt}

\begin{tabular}{@{}p{1.35cm}p{1.35cm}ccrrrr>{\centering\arraybackslash}p{5.8cm}@{}}
\toprule

Benchmark & Instance ID & Panel/Gold & \makecell{Vote\\agreement}
& $z^U$ & $z^C$ & $z^S$ & SCUR & Interpretation / action \\
\midrule

Chatbot Arena & 1110836523 & A/B & 100\%
& .652 & .482 & 1.000 & .967 &
Extreme perturbation sensitivity despite unanimity.\newline
\textbf{Review:} unstable across conditions. \\

HaluEval & 6411 & B/A & 83.3\%
& .999 & .988 & -- & .996 &
High preference ambiguity and cross-entropy.\newline
\textbf{Review:} strong risk evidence. \\

LLMBar & 61 & B/A & 60\%
& .841 & .778 & .754 & .794 &
Multiple moderately elevated signals.\newline
\textbf{Review:} combined risk evidence. \\

LLMEval\textsuperscript{2} & 22 & A/B & 80\%
& .983 & .967 & .833 & .955 &
All three risk signals are elevated.\newline
\textbf{Review:} high risk despite agreement. \\

MMLU & 19396 & A/A & 94.1\%
& .001 & .000 & -- & .001 &
Low preference ambiguity and cross-entropy.\newline
\textbf{Accept:} low review priority. \\

\bottomrule
\end{tabular}
\end{table*}

\section{Conclusion}
SCUR is a label-free risk score developed for assessing individual LLM judge-panel decisions. A cross-fitted latent measurement model provides three risk signals: preference uncertainty, measurement cross-entropy, and perturbation sensitivity. Across five benchmarks, SCUR improves error ranking and selective acceptance, identifies errors that remain difficult to detect under high panel consensus, and is robust to different aggregation rules. These signals further provide interpretable evidence for why individual decisions may warrant human review. Together, our results show that agreement alone does not guarantee decision reliability and that SCUR provides a valuable tool for prioritizing decisions for automatic acceptance or human review.

\section*{AI use statement}

Generative AI tools were used to provide feedback on research methodology and
experimental design, assist with implementation and debugging of experimental
code, support interpretation and presentation of results, and draft and edit
manuscript text for clarity and readability. In addition, LLMs were used as
judge models in the experiments for the LLMBar and
LLMEval\textsuperscript{2} benchmarks, as described in the paper. All
AI-assisted research decisions and manuscript text were reviewed by the
authors, and AI-assisted code and analyses were verified by the authors. The
authors take full responsibility for the methodology, experiments, results,
claims, and final content of the paper.

\section*{Ethics statement}

SCUR is intended to support, rather than replace, human judgment. A low SCUR
score should not be interpreted as a guarantee of correctness, particularly
when multiple LLM judges share systematic errors or biases. The framework
assesses relative decision risk rather than establishing factual correctness.
In high-stakes settings, its scores should be used as evidence for prioritizing
human review rather than as a basis for fully automated decisions.

% \section*{Reproducibility Statement}
% The code and data used in our experiments are available at \url{https://anonymous.4open.science/r/scur-reproduction-68B0/}.

\bibliography{refs}
\bibliographystyle{iclr2027_conference}

\appendix
\section{Model Structure and Estimation}
\label{app:model-structure}

This appendix specifies the latent measurement model and the estimator used
for the reported experiments. The latent item margin describes a preference
shared by the judge panel. Truth labels are used only in evaluation.

\subsection{Measurement Model and Parameter Interpretation}
The measurement
model in Equation~\ref{eq:soft-hlm} is
\begin{equation*}
\operatorname{logit}(p^{\mathrm{llm}}_{ijc}) =a_jq_i+b_j+\delta_c+\kappa_{jc}+\epsilon_{ijc},
 \qquad a_j>0,\qquad \epsilon_{ijc}\sim\mathcal N(0,\sigma_j^2).
\end{equation*}
The model is fitted to log-odds. Judge verdicts are retained for
majority voting and the vote-based baselines. Each non-original condition is
a categorical intervention; Appendix~\ref{app:perturbations} lists the
conditions available for each benchmark.

\begin{table}[htbp]
\caption{Parameters in the fitted continuous-response measurement model.}
\label{tab:model_parameters}
\centering\small
\begin{tabular}{lp{10.6cm}}
\toprule
Parameter & Interpretation \\
\midrule
$q_i$ & Direction and strength of the shared item preference. \\
$a_j$ & Positive discrimination: how strongly judge $j$ responds to $q_i$. \\
$b_j$ & Judge-specific tendency toward canonical candidate $A$. \\
$\delta_c$ & Shared shift under condition $c$, relative to the original condition. \\
$\kappa_{jc}$ & Judge-specific deviation from the shared condition shift. \\
$\sigma_j$ & Residual log-odds scale for judge $j$, estimated after fitting the mean structure. \\
\bottomrule
\end{tabular}
\end{table}

When only original-condition responses
are available, $\delta_c$ and $\kappa_{jc}$ are absent and the model reduces
to $\operatorname{logit}(p^{\mathrm{llm}}_{ij0})=a_jq_i+b_j+\epsilon_{ij0}$.

\subsection{Shrinkage and Coordinate Conventions}
The released estimator uses ridge least-squares updates for the
mean structure. Its blockwise quadratic penalties are
\begin{equation*}
\begin{split}
\mathcal R={}&\frac{\lambda_q}{2}\sum_i q_i^2
 +\frac{\lambda_a}{2}\sum_j(a_j-1)^2
 +\frac{\lambda_b}{2}\sum_j b_j^2\\
 &+\frac{\lambda_\delta}{2}\sum_{c\ne0}\delta_c^2
 +\frac{\lambda_\kappa}{2}\sum_j\sum_{c\ne0}\kappa_{jc}^2.
\end{split}
\end{equation*}
The strengths are $\lambda_q=\lambda_a=\lambda_b=1$,
$\lambda_\delta=5$, and $\lambda_\kappa=10$.
These penalties correspond to Gaussian shrinkage toward zero for
$q,b,\delta,\kappa$ and toward one for $a$, with positivity enforced for
$a$. The model separates shared item and condition components from
judge-specific responses and condition deviations.

The original condition is fixed by $\delta_0=\kappa_{j0}=0$. To fix the
multiplicative scale, after each discrimination update the implementation
sets $g=\exp[J^{-1}\sum_j\log a_j]$ and replaces
$a_j\leftarrow a_j/g$ and $q_i\leftarrow gq_i$, preserving $a_jq_i$.
For each non-original condition, it transfers the mean of
$\kappa_{\cdot c}$ into $\delta_c$, so that $\sum_j\kappa_{jc}=0$ while
preserving the total condition effect. 

\subsection{Optimization and Held-Out Item Inference}
Let $m_{ijc} = \operatorname{logit}(p^{\mathrm{llm}}_{ijc})$. 
Each update regresses the residual left by the other blocks on the current
block. For example, with $\mathcal O_i$ denoting the observed judge--condition
cells for a fit item,
\begin{equation*}
q_i\leftarrow
\frac{\sum_{(j,c)\in\mathcal O_i}a_j
 (m_{ijc}-b_j-\delta_c-\kappa_{jc})}
 {\lambda_q+\sum_{(j,c)\in\mathcal O_i}a_j^2}.
\end{equation*}
Before scale normalization, each discrimination update is clipped to
$[0.05,20]$. The mean-structure fit uses equal observation weights.
After fitting, the residual scale is estimated as
\[
\widehat\sigma_j^2=\frac{1}{n_j}\sum_{(i,c)\in\mathcal O_j}
 (m_{ijc}-\widehat a_j\widehat q_i-\widehat b_j
 -\widehat\delta_c-\widehat\kappa_{jc})^2+10^{-9}.
\]
These residual estimates do not reweight the fitted mean structure. Here $\mathcal O_j$ contains the observed item--condition cells for judge $j$, and $n_j=|\mathcal O_j|$.

We use three random restarts, a nominal 700 iterations, and an
automatic extension up to 2,100 iterations. The standard stopping rule
requires the root mean square change in predicted margins to be below
$10^{-6}$, the maximum change below $10^{-5}$, and the relative change in
the objective below $10^{-7}$ for three successive iterations.
Among converged restarts, the one with the smallest fit objective
is retained. The criterion is mean squared residual plus the
implementation's item- and judge-bias ridge diagnostic; it is used for
convergence and restart selection.

For each seed, items are randomly partitioned into fit, reference, and evaluation splits in a 50/30/20 ratio. Only fit items estimate the
judge and perturbation condition parameters. For a reference or evaluation item, the
leave-one-judge-out margin is obtained using all available conditions from
judges other than $j$:
\begin{equation*}
q_i^{(-j)}=
\frac{\sum_{(\ell,c)\in\mathcal O_i:\ell\ne j}\widehat a_\ell
 (m_{i\ell c}-\widehat b_\ell-\widehat\delta_c-\widehat\kappa_{\ell c})}
 {\lambda_q+\sum_{(\ell,c)\in\mathcal O_i:\ell\ne j}\widehat a_\ell^2}.
\end{equation*}
The corresponding held-out prediction is
$\widehat p_{ijc}=\sigma(\widehat a_jq_i^{(-j)}+\widehat b_j+
\widehat\delta_c+\widehat\kappa_{jc})$. Excluding every condition of the
held-out judge prevents its own responses from determining the item margin
used to assess them. Reference items supply the signal ECDFs, which are
frozen before evaluation. Neither model fitting nor reference normalization
uses ground-truth labels. For numerical stability, the score implementation floors each complement $1-z_i^k$ at $10^{-12}$. A sensitivity component that is absent or constant on the reference split is omitted; the maximum uses the remaining sensitivity component, if any.

\section{Risk Baselines}
\label{app:risk-baselines}

We compare SCUR with vote entropy (VE), predictive entropy (PE),
Dawid--Skene posterior entropy (DS entropy), and mean per-judge MSP
uncertainty. All four use original-condition responses ($c=0$), with
larger scores indicating greater uncertainty.

\subsection{Common Inputs and Evaluation Protocol}
We use the notation of the main text: $J_i$ is the number of available
judges for item $i$, and $p^{\mathrm{llm}}_{ij0}$ is judge $j$'s probability
that the original candidate $A_i$ wins. Let $y_{ij0}\in\{0,1\}$ denote
its binary verdict for that same candidate, with one indicating
$A_i$. The sums below run over the $J_i$ available judges.

We use the binary entropy function \(H\) defined in Section~3.2, with base-2 logarithms for the entropy baselines. 
PE, VE, and MSP require no fitted parameters. DS is fitted on the unlabeled
$\mathcal D_{\mathrm{fit}}$ and then held fixed on
$\mathcal D_{\mathrm{eval}}$. Gold labels are used only for evaluation.

\subsection{Predictive Entropy}
Predictive entropy of verdict-token probabilities has been used to guide
abstention by LLM judges \citep{badshah2026judge}. We apply it to the
panel by averaging the LLM probabilities before taking entropy:
\begin{equation*}
\overline p_i^{\mathrm{llm}}
=\frac{1}{J_i}\sum_{j=1}^{J_i}p^{\mathrm{llm}}_{ij0},
\qquad
\mathrm{PE}_i=H(\overline p_i^{\mathrm{llm}}).
\end{equation*}
The superscript distinguishes this mean from $\overline p_i$ in the main
text, which averages cross-fitted LMM predictions and is used to compute
$U_i$. PE uses the LLM probabilities directly, without fitting the LMM.

PE measures uncertainty in the panel-mean preference. It is largest when
the mean is $0.5$ and small when the mean is near zero or one. A mean near
$0.5$ can reflect individually uncertain judges or confident judges
favoring opposite answers. Its range is $[0,1]$.

\subsection{Vote Entropy}
VE uses the binary verdicts rather than their probabilities:
\begin{equation*}
\mathrm{VE}_i
= H\!\left(\frac{1}{J_i}\sum_{j=1}^{J_i}y_{ij0}\right).
\end{equation*}

It is zero for unanimous votes and largest for an evenly divided panel.
Items with the same vote split have the same VE, even when their LLM
probabilities differ. Thus both unanimously correct and unanimously
incorrect panels receive zero VE. Its range is $[0,1]$.

Vote entropy is a committee-disagreement measure
\citep{dagan1995committee,settles2009active}. Recent LLM-judge calibration
work evaluates majority confidence across repeated verdicts
\citep{radharapu2025calibrating}. For binary verdicts, VE is a monotonic
transformation of one minus majority confidence.
% ere the votes come
% from different judges rather than repeated generations of one judge.

\subsection{Dawid--Skene Posterior Entropy}
A Bayesian variant of the Dawid--Skene model \citep{dawid1979maximum} has been applied to
LLM-evaluator calibration \citep{gao2024bayesian}, while DS is used as an aggregation
baseline in CARE \citep{zhao2026care}. We use entropy of its posterior
as a risk score, extending these aggregation applications to error ranking.

DS estimates a class prior and a confusion matrix for each judge from
original-condition binary verdicts, assuming judges are conditionally
independent given an item's unobserved label. Let $\widehat\rho_i$ be its
fitted posterior probability that candidate $A_i$ is preferred, given
$y_{i10},\ldots,y_{iJ_i0}$. Then
\begin{equation*}
\mathrm{DS\ entropy}_i=H(\widehat\rho_i).
\end{equation*}
This DS posterior is distinct from both the observed LLM probabilities
$p^{\mathrm{llm}}_{ij0}$ and the cross-fitted LMM predictions
$\widehat p_{ij0}$.

We fit the prior and confusion matrices by expectation--maximization on
$\mathcal D_{\mathrm{fit}}$, with additive smoothing of $0.5$.
Initialization uses the proportion of votes for $A_i$, clipped to
$[10^{-3},1-10^{-3}]$. Estimation stops after 200 iterations or when the
maximum change in posterior probabilities is below $10^{-7}$.
The label orientation is chosen by agreement with fit-set majority votes,
without gold labels; swapping the orientation does not change entropy.
The fitted parameters remain fixed for evaluation.

DS entropy measures uncertainty after accounting for learned judge
confusion patterns. It uses binary votes, not $p^{\mathrm{llm}}_{ij0}$. A confidently wrong posterior can still have low entropy.
Its range is $[0,1]$. As in the main text, DS entropy ranks items by risk;
DS aggregation instead uses the posterior to choose their final verdicts.

\subsection{Mean Per-Judge MSP Uncertainty}
Maximum softmax probability (MSP) is a confidence measure
\citep{hendrycks2017baseline}, also evaluated for selective prediction
with LLM judges \citep{radharapu2025calibrating}. Our risk baseline averages
one minus this confidence across judges:
\begin{equation*}
\mathrm{MSP}_i
=\frac{1}{J_i}\sum_{j=1}^{J_i}
\left[1-\max\!\left(p^{\mathrm{llm}}_{ij0},1-p^{\mathrm{llm}}_{ij0}\right)\right].
\end{equation*}
The MSP baseline reports uncertainty: larger
values indicate more uncertain judges. Its range is $[0,0.5]$.
It remains low when every judge is confident, including when confident
judges disagree. The maximum is taken separately for each judge
\emph{before} averaging. Taking
$1-\max(\overline p_i^{\mathrm{llm}},1-\overline p_i^{\mathrm{llm}})$
instead would measure uncertainty of the panel mean and produce the same
ranking as binary PE.

\paragraph{Comparability and ties.}
All four baselines are evaluated against the same $e_i^{\mathrm{MV}}$
defined in the main text, on the same held-out items. AUROC uses average
ranks for equal scores. For $1-\mathrm{AURC}$ and Err@50, we average over
uniform random orderings within tied scores, so file order cannot affect
selective acceptance results. Different raw score ranges do not affect
these ranking-based comparisons.

\section{Benchmark Details and Judging Protocol}
\label{app:real-benchmarks}

\subsection{Benchmark Construction and Validation Labels}
\paragraph{Chatbot-Arena.}
A pairwise-preference benchmark contains $17{,}325$ items,
17 judge models, and $589{,}050$ judge observations. Each seed has
$5{,}198$ reference items and $3{,}465$ evaluation items. The panel-error
target compares the original-condition majority decision with the human
pairwise-preference label.

\paragraph{HaluEval.}
A hallucination-detection benchmark contains $20{,}000$ items,
6 judges, and $120{,}000$ observations. Each seed has $6{,}000$ reference
items and $4{,}000$ evaluation items. The benchmark label determines whether
the original-condition panel verdict is correct.

\paragraph{MMLU.}
A multiple-choice benchmark contains $28{,}084$ items,
17 judges, and $477{,}428$ observations. Each seed has $8{,}425$ reference
items and $5{,}617$ evaluation items. The objective answer key supplies the
validation label.

\paragraph{LLMBar.}

An instruction-following benchmark contains 419 items, 5 judges,
and $12{,}570$ observations across 6 conditions. Each seed has 126 reference
items and 83 evaluation items. The expert preference label determines whether
the original-condition panel verdict is correct.

\paragraph{LLMEval\textsuperscript{2}.}

We use a 200-item subset of the pairwise-preference benchmark, evaluated by 5 judges, yielding
$6{,}000$ observations across 6 conditions. Each seed has 60 reference
items and 40 evaluation items. The source preference label determines whether
the original-condition panel verdict is correct.

\subsection{Perturbations Available for Each Benchmark}
\label{app:perturbations}
\label{app}
A perturbation evaluates the same task and candidate responses under a
modified answer order or judging criterion. Table~\ref{tab:benchmark-perturbations}
summarizes the available conditions.

\begin{table}[htbp]
\caption{Judging conditions and availability of perturbation sensitivity.}
\label{tab:benchmark-perturbations}
\centering\small
\begin{tabular}{l>{\raggedright\arraybackslash}p{8.4cm}c}
\toprule
Benchmark & Conditions & $S$ available? \\
\midrule
Chatbot-Arena & Original and swapped answer orders. & Yes \\
HaluEval / MMLU & Original condition only. & No \\
LLMBar & Original/swapped answer order crossed with standard/strict/evidence-first rubric. & Yes \\
LLMEval\textsuperscript{2} & The same two-order, three-rubric design. & Yes \\
\bottomrule
\end{tabular}
\end{table}

\paragraph{Answer-order swap.}
The two candidate responses exchange their displayed positions while the
task and rubric remain fixed. The resulting probabilities and verdict votes
are expressed relative to the original candidate $A_i$: after a swap, a
probability $p$ of choosing displayed Response A corresponds to probability
$1-p$ of choosing $A_i$.

\paragraph{Rubric changes and paired comparisons.}
For LLMBar and LLMEval\textsuperscript{2}, each answer order is crossed with
three rubrics, described below, giving six conditions. Original order
with the standard rubric defines $c=0$. Each of the other five conditions
is paired with this original condition for the same item and judge;
``paired'' thus refers to comparing judgments of the same candidate pair,
not different items. These comparisons include order changes, rubric
changes, and their combinations. The task and candidate texts remain
fixed. 
% Rubric sensitivity measures dependence on the judging criterion;
% a criterion change need not preserve the preferred answer.

\subsection{Judge Panel}
LLMBar and LLMEval\textsuperscript{2} use the same five judge models:
\begin{itemize}[nosep]
\item \texttt{olmo3-7b-instruct}
\item \texttt{qwen3-30b-a3b-instruct-2507}
\item \texttt{llama4-scout-17b}
\item \texttt{granite41-30b}
\item \texttt{llama3-groq-70b-tool-use}
\end{itemize}
Every judge evaluates every item under all six conditions, yielding 30
judgments per item. Each judgment uses the shared system and task prompts
below, with temperature zero and a one-token A/B output. The probability
$p^{\mathrm{llm}}_{ijc}$ is obtained from verdict-token log-probabilities
and expressed relative to the original candidate $A_i$, as in the main text.

\paragraph{Judging criteria.}
The \emph{standard} rubric asks the judge to choose the better or correct
response. The \emph{strict} rubric makes any material factual or
instruction-following error grounds for rejecting a response. The
\emph{evidence-first} rubric asks the judge to check factual support and
instruction compliance internally before choosing. All three return only
an A/B verdict. The rubric texts appear in the examples below.

\subsection{Prompts and Perturbation Examples}
The system message is identical for both benchmarks and all conditions:
\begin{Verbatim}
You are an impartial evaluator. Return exactly one character: A or B. Do not explain.
\end{Verbatim}

Each box below contains a complete task prompt. We illustrate the three perturbations with three different benchmark
items, showing each perturbed prompt alongside its own original prompt.
Within each pair, the task and candidate texts are unchanged. The original
condition always uses the original answer order and the standard rubric.
Order swaps can also be combined with either modified rubric, giving the
six conditions described above.

\Needspace{0.35\textheight}
\paragraph{Answer-order swap.}
The task asks for two rhymes for ``Moon.'' The original displays ``Soon,
Loon'' as Response A; the swap displays it as Response B. Only the response
positions change, while the task, standard criterion, and displayed-label
semantics remain fixed.

\noindent\textbf{Original.}
\begin{Verbatim}
Evaluation task:
Generate two rhymes for the following word. Moon

Response A:
Soon, Loon

Response B:
Clone, June, Nore, Tune

Label semantics: A means Response A is better; B means Response B is better.

Criterion:
Use the standard criterion: choose the better/correct option.

Answer with A or B only.
\end{Verbatim}

\Needspace{0.32\textheight}
\noindent\textbf{Swapped answer order.}
\begin{Verbatim}
Evaluation task:
Generate two rhymes for the following word. Moon

Response A:
Clone, June, Nore, Tune

Response B:
Soon, Loon

Label semantics: A means Response A is better; B means Response B is better.

Criterion:
Use the standard criterion: choose the better/correct option.

Answer with A or B only.
\end{Verbatim}

\Needspace{0.38\textheight}
\paragraph{Strict rubric.}
This task asks for an eight-character password. The candidates contain ten
and eight characters, respectively. The strict condition replaces the
standard criterion with an explicit instruction to reject material factual
or instruction-following errors; it leaves the task, candidates, and their
order unchanged.

\noindent\textbf{Original.}
\begin{Verbatim}
Evaluation task:
Generate a computer password that is 8 characters long

Response A:
hx3eLXPaPu

Response B:
X7#hklzy

Label semantics: A means Response A is better; B means Response B is better.

Criterion:
Use the standard criterion: choose the better/correct option.

Answer with A or B only.
\end{Verbatim}

\Needspace{0.35\textheight}
\noindent\textbf{Strict rubric.}
\begin{Verbatim}
Evaluation task:
Generate a computer password that is 8 characters long

Response A:
hx3eLXPaPu

Response B:
X7#hklzy

Label semantics: A means Response A is better; B means Response B is better.

Criterion:
Apply the criterion strictly. A response with any material factual or instruction-following error should lose.

Answer with A or B only.
\end{Verbatim}

\Needspace{0.42\textheight}
\paragraph{Evidence-first rubric.}
The task asks for a description of structured restaurant data specifying a
cheap price range. One candidate also calls the restaurant expensive.
The evidence-first condition replaces only the criterion, asking the judge
to check factual support and instruction compliance internally before
choosing. No external evidence is supplied and no written explanation is
requested; the output remains a single A/B verdict.

\noindent\textbf{Original.}
\begin{Verbatim}
Evaluation task:
Describe the given structural data:

inform(name='X',type=placetoeat,eattype=restaurant,pricerange=cheap,near='X')

Response A:
X is a cheap restaurant in the expensive price range near X .

Response B:
X is a cheap restaurant near X .

Label semantics: A means Response A is better; B means Response B is better.

Criterion:
Use the standard criterion: choose the better/correct option.

Answer with A or B only.
\end{Verbatim}

\Needspace{0.38\textheight}
\noindent\textbf{Evidence-first rubric.}
\begin{Verbatim}
Evaluation task:
Describe the given structural data:

inform(name='X',type=placetoeat,eattype=restaurant,pricerange=cheap,near='X')

Response A:
X is a cheap restaurant in the expensive price range near X .

Response B:
X is a cheap restaurant near X .

Label semantics: A means Response A is better; B means Response B is better.

Criterion:
First check factual support and instruction compliance internally, then choose the better/correct option.

Answer with A or B only.
\end{Verbatim}

% \clearpage
\section{High-Consensus Analysis}
\label{app:hc-supplement}

\begin{figure}[htbp]
\centering\includegraphics[width=\linewidth]{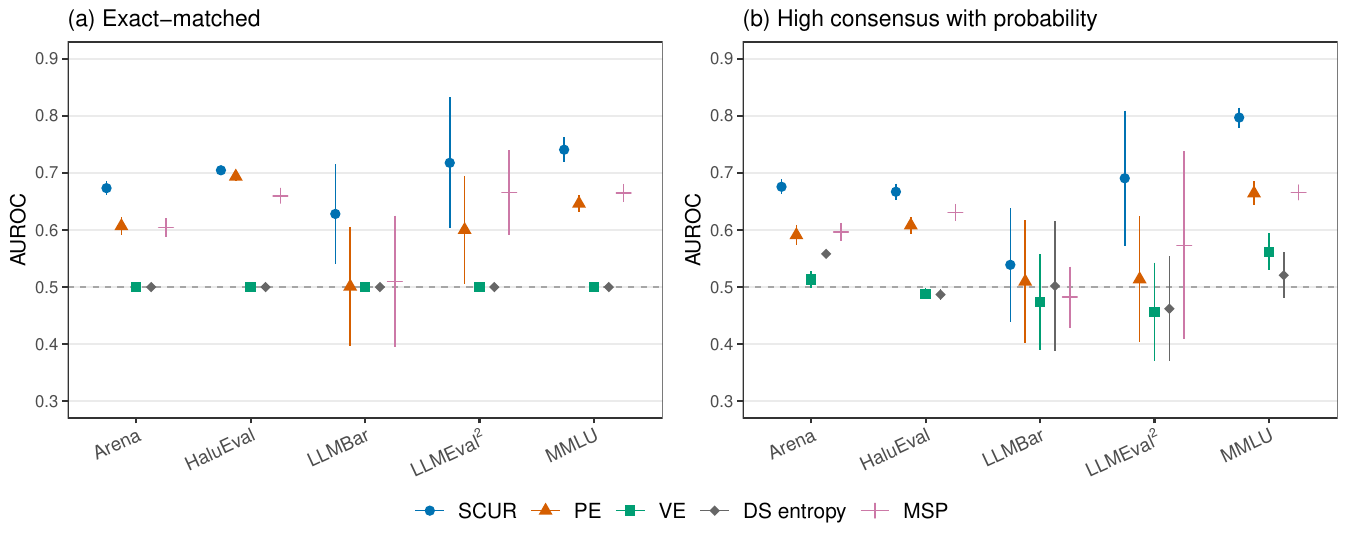}
\caption{\textbf{Error ranking under exact matching and high consensus with probability.}
(a) AUROC comparing items with the same vote pattern; VE and DS entropy
are .5. (b) AUROC among items with panel-mean confidence at least .8.
Both panels compare SCUR with the same four baselines. Points and whiskers
show means and $\pm1$ sample standard deviation over five splits. Dashed lines mark chance performance.}
\label{fig:exact-pattern-control}
\end{figure}

Figure~\ref{fig:exact-pattern-control} complements the high-consensus risk analysis
in Figure~\ref{fig:blindspot-exact-pattern} in two settings: comparing items
with the same vote pattern, and evaluating items with high panel-mean confidence. 
% Both
% analyses use the same judge panels, 50/30/20 splits, and SCUR scores as the
% main evaluation, with majority-vote error as the positive class. Each point
% summarizes over five seeds.

\paragraph{Exact-matched analysis.}
We compare items with the same vote pattern under the original condition:
each judge chooses the same candidate across the compared items. This
keeps both the vote split and the judges supporting each candidate fixed.
We compute AUROC within each group containing both incorrect and correct
decisions, then average these values using the number of incorrect--correct
item pairs in each group as weights:
\[
\mathrm{AUROC_{pattern}}=
\frac{\sum_g n_g^+n_g^-\,\mathrm{AUROC}_g}{\sum_g n_g^+n_g^-},
\]
where $n_g^+$ and $n_g^-$ count incorrect and correct decisions in group $g$.
VE and fitted DS entropy are constant within each group, giving AUROC .5.
SCUR's average AUROC is .693, versus .610 for PE and .621 for MSP
(panel a). Thus SCUR can distinguish incorrect from correct decisions
even when the judges make the same choices.

\paragraph{High consensus with probability.}
Using the panel-mean probability $\overline p_i^{\mathrm{llm}}$ defined in
Appendix~\ref{app:risk-baselines}, high consensus with probability selects
items satisfying
$\max(\overline p_i^{\mathrm{llm}},1-\overline p_i^{\mathrm{llm}})\ge .8$.
This is equivalent to $\mathrm{PE}_i\le H(.8)$: the test asks whether
errors remain distinguishable within a low-PE subset. 
% The filter uses
% only original-condition responses and no gold labels; the risk scores
% are unchanged. Unlike the exact-matched analysis, this subset can contain different
% vote patterns, so VE and DS entropy need not be constant.

SCUR has the highest mean AUROC among the five methods on all benchmarks
in panel (b). On Arena, its AUROC is .676, compared with .596 for the best
baseline, MSP. This supports detecting residual risk even when the panel's
mean probability strongly favors one answer. The LLMBar AUROC remains
near chance (.539), and LLMEval\textsuperscript{2} has only 3--5 errors per
split, limiting the precision of its estimate. The two panels examine complementary aspects of high consensus, showing that strong judge agreement does not eliminate meaningful variation in error risk.

\end{document}